\documentclass[10 pt, conference]{ieeeconf}  
\let\labelindent\relax     
\usepackage{enumitem}
\usepackage{amssymb}
\usepackage{multirow}
\usepackage{graphicx}
\usepackage{pifont}
\usepackage{booktabs}
\usepackage{tabularx}
\usepackage{float}
{\setlength\doublerulesep{0.4pt}}   

\usepackage{cite}[short&compress]
\usepackage{amsmath}
\usepackage{xcolor}

\IEEEoverridecommandlockouts         
 \usepackage{hyperref}

\title{\LARGE \bf
Are We Generalizing from the Exception? An In-the-Wild Study on Group-Sensitive Conversation Design in Human-Agent Interactions}

\author{Ana Müller$^{1}$, Sabina Jeschke$^{2}$ and Anja Richert$^{1}$
\thanks{*The authors acknowledge the financial support by the Federal Ministry of Research, Technology and Space, Germany in the framework FH-Kooperativ 2-2019 (project number 13FH504KX9).}
\thanks{$^{1}$Ana Müller and Anja Richert are with the Cologne Cobots Lab, TH Köln - University of Applied Sciences, 50679 Cologne, Germany {\tt\small ana.mueller@th-koeln.de} $^{2}$ Sabina Jeschke is with KI Park e.V. \& FAU - Friedrich-Alexander University of Erlangen-Nuremberg, Germany}}

\begin{document}

\maketitle
\thispagestyle{empty}
\pagestyle{empty}

\begin{abstract}
This paper investigates the impact of a group-adaptive conversation design in two socially interactive agents (SIAs) through two real-world studies. Both SIAs – Furhat, a social robot, and MetaHuman, a virtual agent – were equipped with a conversational artificial intelligence (CAI) backend combining hybrid retrieval and generative models. The studies were carried out in an in-the-wild setting with a total of \(\emph{N} = 188\) participants who interacted with the SIAs - in dyads, triads or larger groups – at a German museum. Although the results did not reveal a significant effect of the group-sensitive conversation design on perceived satisfaction, the findings provide valuable insights into the challenges of adapting CAI for multi-party interactions and across different embodiments (robot vs. virtual agent) highlighting the need for multimodal strategies beyond linguistic pluralization. These insights contribute to the fields of Human-Agent Interaction (HAI), Human-Robot Interaction (HRI), and broader Human-Machine Interaction (HMI), providing insights for future research on effective dialogue adaptation in group settings.
\end{abstract}

\section{Motivation and Related Work} \label{MRW}

Conversational artificial intelligence (CAI) is the core technology that enables socially interactive agents (SIAs) to understand and generate human language. These agents - including social robots, chatbots, and virtual agents - rely on multimodal signals (e.g., text, speech) to engage in naturalistic interactions with humans \cite{lugrinIntroductionSociallyInteractive2021}. A fundamental component of SIAs is the design of conversations, which structures natural language understanding and -generation into coherent dialogues \cite{kohneConversationDesign2020}. However, while significant progress has been made in the design of dyadic (i.e., one-on-one) interactions, multi-party interactions (MPI, i.e., group interactions) remain an emerging research area \cite{gilletMultipartyInteractionHumans2022,seboRobotsGroupsTeams2020,oliveiraHumanRobotInteractionGroups2021,gilletInteractionShapingRoboticsRobots2024}. This is particularly relevant because the challenges in MPIs mirror those in dyadic settings, but are intensified and introduce additional complexity \cite{gilletMultipartyInteractionHumans2022,nigroSocialGroupHumanRobot2025}, negatively impacting the quality of Human-Machine Interaction (HMI) \cite{oliveiraHumanRobotInteractionGroups2021,seboRobotsGroupsTeams2020}. To establish SIAs as effective participants in complex social environments, a deeper understanding of their impact on group dynamics is essential \cite{seboRobotsGroupsTeams2020}. Yet, more empirical work is needed to explore the nuances of group interaction, especially since social interactions frequently occur in group contexts \cite{frauneHumanGroupPresence2019}. These scenarios demand strategies that can accommodate multiple individuals simultaneously, particularly in areas such as turn-taking \cite{skantzeTurntakingConversationalSystems2021,skantzeExploringTurntakingCues2015}, gaze behavior \cite{vazquezRobotAutonomyGroup2017}, open-domain dialogue \cite{addleseeMultipartyConversationalSocial2024}, and speaker diarization \cite{muraliImprovingMultipartyInteractions2023}. As a result, adaptive conversation design has gained increasing attention, driven by advances in generative AI that support more dynamic MPIs \cite{addleseeMultipartyMultimodalConversations2024,bensch2024,muraliImprovingMultipartyInteractions2023}. Recent approaches, such as the ‘Template and Graph-Based Modeling’–framework \cite{gilletTemplatesGraphNeural2025}, propose a structured decision making model that decouples ‘who to address’ from ‘what to communicate’, using graph neural networks for adaptive small group interactions. Moreover, recent progress in large language models (LLMs), such as GPT-4 (OpenAI, USA, \cite{openaiGPT4TechnicalReport2024}) and LLaMA 3 (Meta, USA, \cite{dubeyLlama3Herd2024}), opens new possibilities for group-sensitive CAI. These models can dynamically adjust their responses based on context and potentially differentiate between dyadic and group scenarios using techniques such as automatic speaker diarization \cite{muraliImprovingMultipartyInteractions2023} or vision-based person detection \cite{bensch2024}. Studies by Addlesee et al. \cite{addleseeMultipartyMultimodalConversations2024,addleseeMultipartyConversationalSocial2024} further show how generative AI can enhance MPIs in open dialogues, marking a step toward more flexible, context-aware SIAs.

\begin{figure*} 
    \centering
    \includegraphics[width=\textwidth]{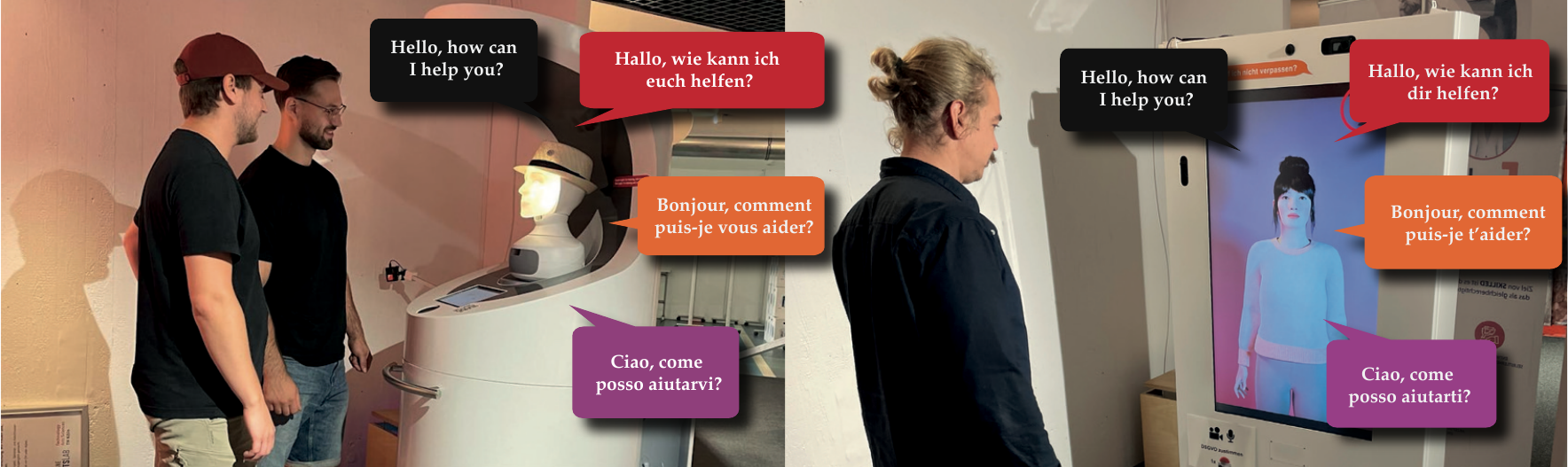}  
    \caption{Interactions at the museum featuring SIAs, both named MIRA: On the left, the Furhat robot engages in group-sensitive conversations. On the right, MetaHuman demonstrates a dyadic conversation design. Examples illustrate distinct conversation strategies across contexts, adapting to German, French, and Italian, in contrast to English, which remains “How can I help you?” throughout. To protect privacy, individuals shown are not actual study participants.}
    \label{int_lang}
\end{figure*}

Beyond ongoing challenges \emph{and} solutions, a key reason for the lack of adequate strategies for MPI is the historical focus of CAI on dyadic interactions. This is evident in commercial voice assistants, such as Alexa (Amazon, USA) and Siri (Apple, USA), which are designed to assist a single user \cite{guzmanMakingAISafe2017,bohusModelsMultipartyEngagement2009}. Consequently, current SIAs struggle to adapt their interaction strategies for MPIs. An overlooked aspect is the linguistic adaptation required for group interactions: Many languages, including German, French, Italian, and Polish, distinguish between singular and plural forms of ‘you’ unlike English, which does not explicitly encode this difference. As a result, conversation designs developed for English have often been applied to other languages without proper adaptation. For example, a German-speaking SIA asking “Wie kann ich \emph{dir} helfen?” (“How can I help you?” [singular]) instead of “Wie kann ich \emph{euch} helfen?” (“How can I help you?” [plural]) can inadvertently exclude group members from the interaction (Fig. \ref{int_lang}) \cite{bensch2024}. This oversight does not stem from AI being trained in English, but is rather a result of transferring English-based design conventions to languages with fundamentally different grammatical rules. Although some exceptions exist (e.g., Turkish, which resembles English in this regard), most conversation design frameworks are still rooted in dyadic interaction assumptions, limiting their applicability to MPIs.

Building on these observations, the present studies evaluate whether a group-sensitive conversation design improves interaction quality in MPIs, and whether the embodiment of the SIA (robot vs. virtual agent) influences the perception in this regard. We test the following hypotheses:

\begin{itemize}
    \item \textbf{H1:} A group-sensitive conversation design improves the perceived quality of the conversation in MPIs.
    \item \textbf{H2:} Differences in embodiment (robot vs. virtual agent) influence user satisfaction in group interactions.
\end{itemize}

These hypotheses are examined through two independent real-world studies, each testing a different embodiment of SIAs with a shared backend. In previous work \cite{bensch2024}, we introduced a group-adaptive conversation design framework for SIAs, integrating computer vision to distinguish dyadic from group interactions, and a hybrid rule-based and LLM-driven backend to generate contextually adaptive responses. This framework was specifically developed to overcome the limitations of English-centric design when applied to languages with pluralization distinctions. The present study extends this work through empirical evaluation in two real-world deployments at the Deutsches Museum Bonn, involving a Furhat robot (Furhat Robotics, Sweden); \(\emph{n} = 113\) and a MetaHuman virtual agent developed using Unreal Engine (Epic Games, Cary, USA); \(\emph{n} = 75\). In contrast to previous research, often limited to dyads or controlled Wizard-of-Oz setups \cite{jungRobotsWildTime2018}, this study provides a comparative evaluation of autonomous SIAs in naturalistic, unscripted HMIs. To the best of our knowledge, this is the first real-world validation of whether adaptive pluralization strategies improve satisfaction in MPIs evaluated with two distinct embodiments of SIAs. The findings therefore contribute to the fields of Human-Agent Interaction (HAI), Human-Robot Interaction (HRI), and broader Human-Machine Interaction (HMI).

\section{Materials and Methods}

The Deutsches Museum Bonn (DMB), a museum of science and technology in Germany, served as the testing site for two studies: one involving the Furhat robot (\emph{n} = \(113\)) and the other with the MetaHuman virtual agent (\emph{n} = \(75\)). Figure \ref{int_lang} illustrates interactions with both SIAs at the DMB. The SIAs were chosen to determine whether the group-adaptive conversation design functions consistently in both embodiments – physical robots and virtual agents – supporting cross-platform validation. Participants engaged in natural interactions where the SIAs employed either a dyadic or group-sensitive conversation design. The Furhat study was conducted from July 23\textsuperscript{rd} to August 15\textsuperscript{th}, 2024, and the MetaHuman study took place from October 18\textsuperscript{th} to November 4\textsuperscript{th}, 2024, both corresponding to the museum's opening hours (Tuesday-Sunday). The SIAs were placed near the entrance in an accessible but semi-enclosed exhibition area. To comply with data protection regulations, each system remained idle until visitors gave their GDPR-compliant consent by pressing a buzzer, which then activated the interaction. The research activities were approved by the Ethics Research Committee of TH Köln (application no. THK-2023-0004).

\subsection{Research System}

The research system is based on the modular architecture for group-adaptive SIAs, as described in \cite{bensch2024}. Based on this framework, we distinguish between two layers of design: \textit{group-adaptive conversation design}, which refers to the technical capability of SIAs to dynamically adjust responses, and \textit{group-sensitive design}, which denotes surface-level behaviors perceived by users in group settings. The architecture of the MetaHuman agent is documented in detail in \cite{chojnowskiHumanlikeNonverbalBehavior2025a}. The core back-end functionality of both SIAs was identical or built on closely related architectures, with only minor adjustments. Both systems were connected to the same retrieval-based CAI, enabling smooth interaction in both German and English. The knowledge database included curated content on exhibitions, robotics, and small talk. Generative AI was used to expand the training set with various conversational examples (see \cite{bensch2024} for details). For queries falling outside the knowledge base, LLaMA 3 was used as a fallback model, with consistent prompt structures applied in both studies. To distinguish dyadic from group interactions, YOLOv5m \cite{jocherUltralyticsYolov5V312020} was used in the Furhat study \cite{bensch2024}, while MediaPipe EfficientDet-Lite \cite{googleaiedgeMediaPipe} was used in the MetaHuman study \cite{chojnowskiHumanlikeNonverbalBehavior2025a}, adapting to different hardware needs of the SIAs. The functionality of both computer vision models was confirmed in advance through laboratory tests and multi-day field trials. In both cases, frame averaging was implemented to improve detection stability and reduce short-term misclassifications. To comply with GDPR regulations, all data processing was performed locally on a NVIDIA RTX 4090 GPU, ensuring that no personal data were transmitted to external servers.

\subsection{Research Design}

Following independent interactions with the SIAs, museum visitors were asked to complete a digital survey on a tablet using SoSci Survey (Germany). Both studies followed an A/B test structure to assess how group-sensitive conversation design influenced user satisfaction. Although complete randomization was impractical in this in-the-wild setting, we systematically balanced the conditions throughout the days and weekends and holidays to achieve an equitable distribution. Participation by minors was allowed only with the consent of a legal guardian. To accommodate families during school holidays, robot-themed coloring pages generated with DALL\textperiodcentered E (OpenAI, USA) were provided to keep children engaged, allowing adults to complete the survey without distraction. Participants received small project-related gifts, such as RFID blockers, key rings, and bags, as a token of appreciation.

\subsection{Questionnaire Design} \label{QDesign}

In addition to demographic data such as age, gender, and educational background, perceived conversational satisfaction was assessed using 11 items on a 5-point Likert scale (higher values indicate greater agreement). The scale covered perceived competence, likability, linguistic appropriateness, intuitiveness, conversational flow, entertainment value, and appropriateness of responses. Furthermore, perceptions of speech-to-text and text-to-speech quality were included, as well as potential disappointment in the conversation and its perceived error rate (the latter two were reverse-coded). To ensure measurement quality, internal consistency was evaluated. Cronbach's $\alpha$ for conversational satisfaction (11 items) was calculated for the general data set, which revealed good reliability ($\alpha$ = \(.828\), \(\emph{N} = 183\), with \(5\) cases excluded). Scale stability was also examined separately for each SIA, revealing strong agreement in the two studies: $\alpha$ = \(.802\) for Furhat and $\alpha$ = \(.848\) for MetaHuman. This indicates a consistent and reliable measurement of conversational satisfaction between different embodiments of agents. Given the distinct SIA embodiments in each iteration, this comparison allowed an evaluation of scale stability and potential differences in test-retest reliability \cite{oliveiraHumanRobotInteractionGroups2021}.

At the beginning of the questionnaire, participants indicated how many people had interacted with the SIA alongside them, allowing classification of dyadic versus group interactions. A validation check was included to allow participants to revise their reported group size if needed. To further ensure measurement precision, a parallel observation protocol was implemented: researchers recorded the session start and observed group size as ground truth (see Sect.~\ref{datacleansing}). Groups were defined as systematic interaction patterns that involve at least two humans \cite{fineGroupCultureInteraction2012}. Consequently, a group interaction was defined as two or more people engaging with the SIA simultaneously within the designated interaction area, this definition also guided the implementation of the technical framework in \cite{bensch2024}. In the present studies, most MPIs involved pairs; triads were less common, and larger groups were rare. The maximum group size observed was five individuals, including the respondent.

\subsection{Data Cleansing} \label{datacleansing}

Before data cleaning, the dataset included \(\emph{n} = 188\) responses for Furhat and \(\emph{n} = 113\) for MetaHuman. To ensure data quality, all responses were cross-validated against the observation protocol (Sect. \ref{QDesign}). This allowed for the correction of false positives (participants indicating group presence when alone) and false negatives (participants reporting solitude despite being in a group). These corrections ensured an accurate classification of the interaction types, thus strengthening the validity of the A/B test conditions. Several additional exclusion criteria were applied. The responses to the English version of the questionnaire were removed due to the expected uniformity between dyadic and MPIs in English (Sect.~\ref{MRW}). In addition, responses with excessively short completion times, irregular patterns, or missing key information were excluded. After cleaning, the final dataset comprised \(\emph{N} = 188\) valid entries: \(\emph{n} = 113\) for Furhat and \(\emph{n} = 75\) for MetaHuman.

\section{Results and Discussion} \label{CD_deskript}

Following data cleaning, we obtained the distribution of conditions: C-A (dyadic conversation design: Furhat \(\emph{n} = 55\), MetaHuman \(\emph{n} = 39\)) and C-B (group-sensitive conversation design: Furhat \(\emph{n} = 58\), MetaHuman \(\emph{n} = 36\)) were equally distributed across iterations. Distinguishing dyadic interactions (always with dyadic conversation design due to technical design) from group interactions (using dyadic or group-sensitive conversation design) is critical. Although the group conditions were balanced, the sample sizes varied between both studies (\(\emph{n} = 113\) vs. \(\emph{n} = 75\)), which could affect variance analysis and the choice of statistical techniques.

\subsection{Descriptives Results}

In the study with Furhat, participants had a mean age of \(40.31\) years (\(\emph{SD} = 14.81\)), while in the study with MetaHuman the mean age was \(42.12\) years (\(\emph{SD} = 14.51\)), showing a similar age distribution. Men were overrepresented in both iterations (\(70\) female, \(109\) male, \(3\) non-binary / diverse, and 6 participants who did not disclose their gender). The sample included individuals with academic and non-academic backgrounds, as well as students. The majority (\(> 55\) \(\%\)) had at least one university degree, while other educational backgrounds were less common. The sample represented a diverse range of professional backgrounds, including employees, self-employed individuals, students, soldiers, and retirees. This variety may contribute to different expectations when interacting with SIAs. Conversational satisfaction (overall dataset: \(\emph{M} = 3.39\), \(\emph{SD} = .53\)) showed slight variations between Furhat (\emph{M} = 3.46, \emph{SD} = .48) and MetaHuman (\(\emph{M} = 3.27\), \(\emph{SD} = .58\)). However, these differences remain within a small range, suggesting that there is no substantial discrepancy between the SIAs.

\subsection{Inferential Analysis}

The inferential analysis, conducted with IBM SPSS Statistics 30, aimed to examine the impact of conversation design (dyadic vs. group-sensitive) on perceived conversation satisfaction. This section presents inferential statistical analyses based on conversation design and the interaction category separately for the two SIAs. To test for significant differences, analysis of variance (ANOVA) was applied, provided that the assumptions of normal distribution (Shapiro-Wilk test) and homogeneity of variance (Levene's test) were met. ANOVA examines how one or more independent variables affect a dependent variable \cite{fieldDiscoveringStatisticsUsing2024}. In this case, a one-way ANOVA was used to assess the effect of the independent variable ‘interaction category’ on the dependent variable ‘satisfaction’. As each embodiment (Furhat vs. MetaHuman) was tested in an independent study with distinct participant groups, a factorial design (e.g., two-way ANOVA) was not applicable. Therefore, the interaction category was treated as a categorical variable with three levels: dyadic interaction, group interaction with dyadic conversation design, and group interaction with group-sensitive conversation design. ANOVA tests whether differences in the dependent variable, caused by variations in the independent variable, are statistically significant, i.e., whether observed differences are unlikely to occur by chance \cite{fieldDiscoveringStatisticsUsing2024}. The significance level was established at \(5\) \(\%\) ($\alpha$ = \(.05\)), which means that a result is considered statistically significant if the value of \emph{p} is below \(.05\) \cite{fieldDiscoveringStatisticsUsing2024}. The 11-item conversational satisfaction scale was analyzed across the three conditions, with dyadic interaction serving as the baseline condition (dyadic condition, DC).

\subsection{Conversational Satisfaction Analysis - Furhat}

\begin{figure}[htbp]
\centering
\includegraphics[width=1\linewidth]{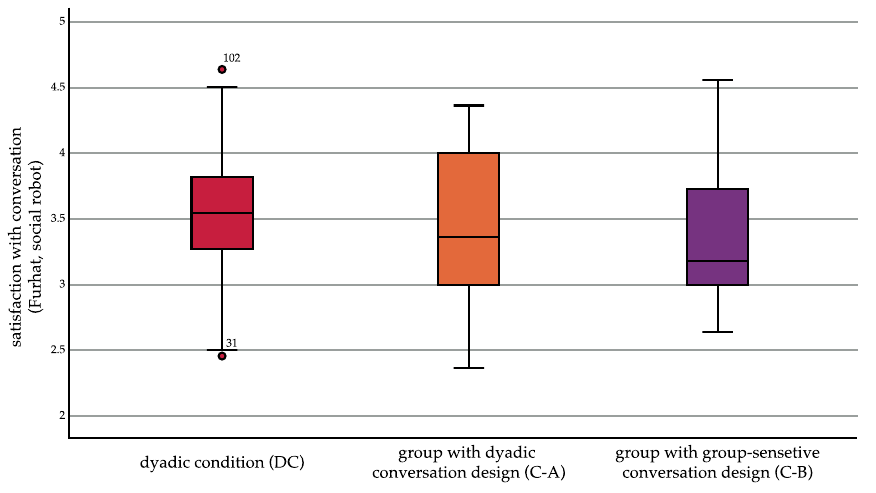}
\caption{Box plot of conversational satisfaction for Furhat, categorized by interaction type. DC exhibited the highest variance with a median around \(3.0\). C-A had a higher median (approximately \(3.5\)), while C-B was similar to C-A but with a narrower distribution.}
\label{Fig_Boxplot_Conv_Furhat}
\end{figure}

Before performing the ANOVA for conversational satisfaction with Furhat, its assumptions were tested. The box plot (Fig. \ref{Fig_Boxplot_Conv_Furhat}) illustrates the distribution of satisfaction between the three conditions. The median (black line in the box plots) varied slightly between conditions. DC had a slightly higher mean satisfaction score than C-B, while C-A showed a similar median to DC. C-B had the lowest median and the variability was highest in C-A, indicating a greater dispersion in satisfaction. Some outliers were observed: The participant \(102\) (DC) reported high satisfaction, while the participant \(31\) (DC) reported low satisfaction. A lower satisfaction trend was observed in group interactions with a group-sensitive conversation design. If statistically significant, this would indicate that the group-sensitive conversation design negatively impacted satisfaction compared to the dyadic conversation design in group settings.

A Levene test confirmed the homogeneity of the variances (\(\emph{p} > .05\)), allowing for ANOVA. The results (Tab. \ref{tab:anova:Furhat:Conv}) did not show significant differences between conditions (\(\emph{F} = 1.96\), \(\emph{p} = .15\), $\eta^2$ \(= .04\)), suggesting that the conversation design did not have a statistically significant impact on satisfaction.

\begin{table}[ht]
\small
\centering
\caption{One-way ANOVA results for Furhat.}
\label{tab:anova:Furhat:Conv}
\begin{tabular}{lcccc}
\toprule
\textbf{Source} & \textbf{\emph{df}} & \textbf{\emph{F}} & \textbf{\emph{p}-Value} & \textbf{Partial \emph{$\eta^2$}} \\
\midrule
Conversation Design Mode & 2 & 1.96 & .15 & .035 \\
\bottomrule
\end{tabular}
\vspace{0.3cm}
$R^2 = 0.035$, adjusted $R^2 = 0.017$.
\end{table}

Tukey and Bonferroni post hoc tests did not show significant differences between the conditions (\(\emph{p} > .05\)). The largest mean difference between DC and C-B was observed, but was not significant (\(\emph{p} = .13\) for Tukey, \(\emph{p} = .16\) for Bonferroni). The findings do not support the hypothesis that a pluralized conversation design with group-sensitiveness improves conversational satisfaction in group interactions. C-B was not significantly more satisfied than C-A. This suggests that pluralization alone may not improve the quality of interaction in group settings with Furhat.

\subsection{Conversational Satisfaction Analysis - MetaHuman}

\begin{figure}[htbp]
\centering
\includegraphics[width=1\linewidth]{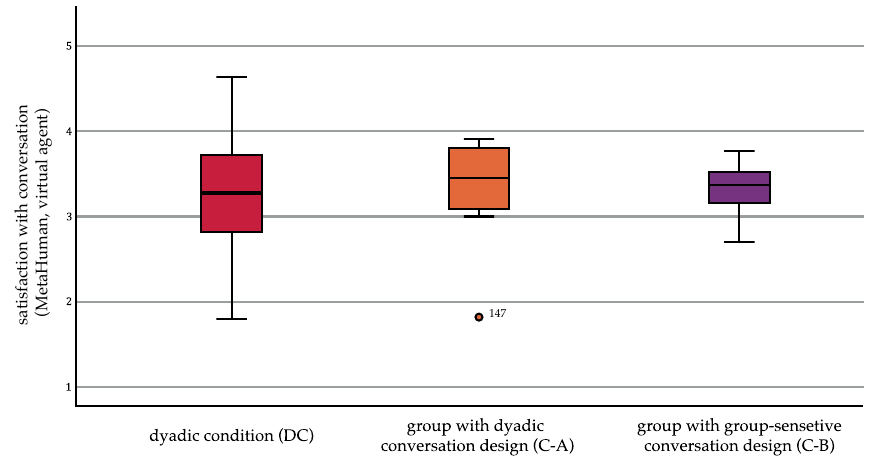}
\caption{Box plot of conversational satisfaction scores for MetaHuman, categorized by interaction type.}
\label{Fig_Boxplot_MH}
\end{figure}

Similar to the Furhat analysis, the conversational satisfaction for MetaHuman was examined, with dyadic interactions serving as the baseline condition (DC). Among the conditions, DC showed the highest variance in satisfaction, with a median around \(3.0\). In contrast, C-A  (group, dyadic) had a slightly higher median of approx. \(3.5\), while C-B (group, group-sensitive) was similar to C-A but exhibited a more concentrated distribution with fewer extreme values (Fig. \ref{Fig_Boxplot_MH}). DC had the highest variance, while C-A and -B showed a tighter distribution with fewer extreme values.	An outlier was identified in C-A (participant \(147\), satisfaction score \(< 2\)). Visually, DC appeared to have lower satisfaction than the group conditions (C-A \& -B).

\begin{table}[htbp]
\small
\centering
\caption{Results of the Shapiro-Wilk test for MetaHuman.}
\label{tab:shapiro_wilk}
\begin{tabular}{lcc}
\toprule
\textbf{Condition} & \textbf{Shapiro-Wilk \emph{p}-Value} & \textbf{Interpretation} \\
\midrule
DC & .89 & Normally distributed \\
C-A & \textcolor{red}{.05} & \textcolor{red}{Not normally distributed} \\
C-B & .73 & Normally distributed \\
\bottomrule
\end{tabular}
\end{table}

To test the assumption of normality, a Shapiro-Wilk test (Tab. \ref{tab:shapiro_wilk}) was performed. The results showed that C-A deviated significantly from normality (\(\emph{p} = .05\)), while DC and C-B were normally distributed. Since C-A violated the normality assumption (\emph{p} \(< .05\)) and the sample sizes were highly uneven, a Kruskal-Wallis test was performed instead of an ANOVA, as this nonparametric test is suitable for small samples and nonnormal distributions \cite{fieldDiscoveringStatisticsUsing2024}. The Kruskal-Wallis test (Tab. \ref{KWCMH}) did not reveal significant differences in conversational satisfaction between conditions (\(\emph{H}(2) = 0.55, \emph{p} = .76\)). Since \(\emph{p} = .76 > .05\), no significant differences were found. Furthermore, post hoc pairwise comparisons (corrected for Bonferroni) confirmed this result, showing no significant differences between any conditions.

\begin{table}[htbp]
\centering
\caption{Kruskal-Wallis test results for MetaHuman.} 
\label{KWCMH}
\begin{tabular}{lc}
\toprule
\textbf{Parameter} & \textbf{Value} \\
\midrule
Total \emph{N} & 71 \\  
Test Statistic (\emph{H}) & .55$^a$ \\
Degrees of Freedom (\emph{df}) & 2 \\
\emph{p}-Value (two-tailed) & .76 \\
\midrule
\multicolumn{2}{l}{\footnotesize $^a$ Adjusted for tied ranks.} \\
\bottomrule
\end{tabular}
\end{table}

The group conditions (C-A \& B) did not differ significantly from the dyadic interactions (DC). This suggests that conversation design, whether dyadic or group-sensitive, had no measurable impact on perceived conversational satisfaction. Although visual differences hinted at lower satisfaction in DC, statistical tests did not confirm this trend. The results for the virtual agent align with those for the social robot, reinforcing the finding that conversation design did not have a significant effect on conversational satisfaction.

\subsection{Power Analysis}

To assess the statistical power of the analysis, a post hoc power analysis was performed using G*Power 3.1 \cite{faulStatisticalPowerAnalyses2009} for the whole sample. The statistical power calculated for one-way ANOVA (\(\alpha = .05\), \(f = .19 \), (\(\emph{N} = 188\)), three groups) was \(.63\). This indicates a probability 63 \% of detecting a true effect, while the risk of a Type II error (failing to detect an existing effect) is \(37\) \(\%\). Since the recommended minimum power is (\(80\) \(\%\)), this result suggests that the sample size may not be sufficient to reliably detect small effects. Power analysis revealed that a minimum of \(270\) participants would have been required to achieve adequate statistical power. The reduced power could explain the nonsignificant effects in the ANOVA, indicating that these results should be carefully interpreted, as they might result from inadequate statistical sensitivity in the in-the-wild setting rather than the absence of an effect. Future studies should consider larger sample sizes to improve the robustness of the findings and the probability of identifying smaller effects.

\section{Conclusion, Limitations \& Future Directions}

This study examined whether a group-sensitive conversation design improves perceived conversation satisfaction in MPI and whether different SIA embodiments (social robot vs. virtual agent) influence these perceptions. The results did not indicate a statistically significant effect of group-sensitive conversation design, suggesting that linguistic pluralization alone may not be sufficient to enhance group interactions in SIAs. However, this finding should not be interpreted as evidence that group-sensitiveness is ineffective in general, but rather that its impact is context-dependent. One possible explanation is that users may not have consciously perceived the adaptation, particularly in short, in-the-wild interactions. Additionally, some interactions started as dyadic and later transitioned into MPI, meaning that pluralization may not always have been required. Although the system correctly applied group-sensitive adaptation when needed, some interactions may have consisted mainly of neutral intents (e.g., factual responses to ‘What is the weather?’), where distinctions were unnecessary. As a result, the adaptation may have occurred without the participants explicitly noticing it.

Another factor could be that users expect SIAs to operate in dyads, even in group interactions. This suggests that the design of group-sensitive conversations needs reinforcement through multimodal strategies, such as adaptive gaze shifts, turn-taking mechanisms, or prosodic adjustments, to more effectively signal group awareness, as indicated by related work, such as \cite{skantzeExploringTurntakingCues2015,gilletTemplatesGraphNeural2025,vazquezRobotAutonomyGroup2017} in conjunction with the design of adaptive group-sensitive conversations. For example, depending on the context of the interaction, humans in groups do not consistently address one group but adapt between dyads and groups depending on the context. This highlights the distinction between adaptive systems, which technically modify their behavior, and group-sensitive experiences, which depend on users perceiving those modifications as socially meaningful in a group context. This has similarities to the framework recently presented by Gillet et al. \cite{gilletTemplatesGraphNeural2025} for navigating small groups, but with linguistic sensitivity, in languages where necessary.

With an achieved power of \(.63\), the study might lack sufficient power to identify small effects. A larger sample size, estimated at \emph{N =} \(270\), would be needed to determine whether the group-sensitive conversation design has no observable impact or if the effect size is too minor to be detected. Future studies with larger samples should include manipulation checks (e.g. 'Did the agent's response seem directed specifically toward you, or was it directed at the group as a whole?’) to evaluate perceptions of adaptation. In addition, a qualitative follow-up study would help determine whether the users were aware of the adaptation.

Although this study did not find a significant effect in this specific context, it provides important insights into the challenges of adapting conversation design for multi-party interactions. The results emphasize the need for a multimodal and context-sensitive approach, rather than relying only on linguistic pluralization. Future work should explore how different adaptive mechanisms, such as gaze behavior, response timing, and social cues, can be combined with group-sensitive conversation design to improve interactions in group settings. Furthermore, it remains an open question whether different interaction contexts, such as elderly care facilities or educational settings, might yield different results due to longer engagement and different user expectations. However, by moving beyond purely linguistic adaptation, this research lays the foundation for more flexible and context-aware SIAs capable of naturally engaging in dynamic group environments.

\section*{ACKNOWLEDGMENT}

We thank our collaboration partner DB Systel GmbH, the Deutsches Museum Bonn and all other collaborators for their assistance and contributions.

\bibliographystyle{IEEEtran}
\bibliography{literature}

\begin{thebibliography}{10}
\providecommand{\url}[1]{#1}
\csname url@samestyle\endcsname
\providecommand{\newblock}{\relax}
\providecommand{\bibinfo}[2]{#2}
\providecommand{\BIBentrySTDinterwordspacing}{\spaceskip=0pt\relax}
\providecommand{\BIBentryALTinterwordstretchfactor}{4}
\providecommand{\BIBentryALTinterwordspacing}{\spaceskip=\fontdimen2\font plus
\BIBentryALTinterwordstretchfactor\fontdimen3\font minus \fontdimen4\font\relax}
\providecommand{\BIBforeignlanguage}[2]{{%
\expandafter\ifx\csname l@#1\endcsname\relax
\typeout{** WARNING: IEEEtran.bst: No hyphenation pattern has been}%
\typeout{** loaded for the language `#1'. Using the pattern for}%
\typeout{** the default language instead.}%
\else
\language=\csname l@#1\endcsname
\fi
#2}}
\providecommand{\BIBdecl}{\relax}
\BIBdecl

\bibitem{lugrinIntroductionSociallyInteractive2021}
\BIBentryALTinterwordspacing
B.~Lugrin, ``Introduction to {{Socially Interactive Agents}},'' in \emph{The {{Handbook}} on {{Socially Interactive Agents}}}, 1st~ed., ser. The {{Handbook}} on {{Socially Interactive Agents}}, B.~Lugrin, C.~Pelachaud, and D.~Traum, Eds.\hskip 1em plus 0.5em minus 0.4em\relax ACM, 2021, no. VOL 1, pp. 1--20. [Online]. Available: \url{https://dl.acm.org/doi/10.1145/3477322.3477324}
\BIBentrySTDinterwordspacing

\bibitem{kohneConversationDesign2020}
A.~Kohne, P.~Kleinmanns, C.~Rolf, and M.~Beck, ``Conversation design,'' in \emph{Chatbots: Aufbau und Anwendungsmöglichkeiten von autonomen Sprachassistenten}, A.~Kohne, P.~Kleinmanns, C.~Rolf, and M.~Beck, Eds.\hskip 1em plus 0.5em minus 0.4em\relax Springer Fachmedien, 2020, pp. 83--97.

\bibitem{gilletMultipartyInteractionHumans2022}
S.~Gillet, M.~Vázquez, C.~Peters, F.~Yang, and I.~Leite, ``Multiparty {{Interaction Between Humans}} and {{Socially Interactive Agents}},'' in \emph{The {{Handbook}} on {{Socially Interactive Agents}}}, 1st~ed., B.~Lugrin, C.~Pelachaud, and D.~Traum, Eds.\hskip 1em plus 0.5em minus 0.4em\relax ACM, 2022, pp. 113--154.

\bibitem{seboRobotsGroupsTeams2020}
S.~Sebo, B.~Stoll, B.~Scassellati, and M.~F. Jung, ``Robots in {{Groups}} and {{Teams}}: {{A Literature Review}},'' \emph{Proceedings of the ACM on Human-Computer Interaction}, vol.~4, pp. 1--36, 2020.

\bibitem{oliveiraHumanRobotInteractionGroups2021}
R.~Oliveira, P.~Arriaga, and A.~Paiva, ``Human-{{Robot Interaction}} in {{Groups}}: {{Methodological}} and {{Research Practices}},'' \emph{Multimodal Technologies and Interaction}, vol.~5, no.~10, p.~59, 2021.

\bibitem{gilletInteractionShapingRoboticsRobots2024}
\BIBentryALTinterwordspacing
S.~Gillet, M.~Vázquez, S.~Andrist, I.~Leite, and S.~Sebo, ``Interaction-{{Shaping Robotics}}: {{Robots That Influence Interactions}} between {{Other Agents}},'' \emph{J. Hum.-Robot Interact.}, vol.~13, no.~1, pp. 12:1--12:23, 2024. [Online]. Available: \url{https://dl.acm.org/doi/10.1145/3643803}
\BIBentrySTDinterwordspacing

\bibitem{nigroSocialGroupHumanRobot2025}
M.~Nigro, E.~Akinrintoyo, N.~Salomons, and M.~Spitale, ``Social {{Group Human-Robot Interaction}}: {{A Scoping Review}} of {{Computational Challenges}},'' in \emph{Proceedings of the 2025 {{ACM}}/{{IEEE International Conference}} on {{Human-Robot Interaction}}}, ser. {{HRI}} '25.\hskip 1em plus 0.5em minus 0.4em\relax IEEE Press, 2025, pp. 468--478.

\bibitem{frauneHumanGroupPresence2019}
M.~R. Fraune, S.~Šabanović, and T.~Kanda, ``Human {{Group Presence}}, {{Group Characteristics}}, and {{Group Norms Affect Human-Robot Interaction}} in {{Naturalistic Settings}},'' \emph{Frontiers in Robotics and AI}, vol.~6, 2019.

\bibitem{skantzeTurntakingConversationalSystems2021}
G.~Skantze, ``Turn-taking in {{Conversational Systems}} and {{Human-Robot Interaction}}: {{A Review}},'' \emph{Computer Speech \& Language}, vol.~67, p. 101178, 2021.

\bibitem{skantzeExploringTurntakingCues2015}
G.~Skantze, M.~Johansson, and J.~Beskow, ``Exploring {{Turn-taking Cues}} in {{Multi-party Human-robot Discussions}} about {{Objects}},'' in \emph{Proceedings of the 2015 {{ACM}} on {{International Conference}} on {{Multimodal Interaction}}}, ser. {{ICMI}} '15.\hskip 1em plus 0.5em minus 0.4em\relax Association for Computing Machinery, 2015, pp. 67--74.

\bibitem{vazquezRobotAutonomyGroup2017}
\BIBentryALTinterwordspacing
M.~Vázquez, E.~J. Carter, B.~McDorman, J.~Forlizzi, A.~Steinfeld, and S.~E. Hudson, ``Towards {{Robot Autonomy}} in {{Group Conversations}}: {{Understanding}} the {{Effects}} of {{Body Orientation}} and {{Gaze}},'' in \emph{Proceedings of the 2017 {{ACM}}/{{IEEE International Conference}} on {{Human-Robot Interaction}}}.\hskip 1em plus 0.5em minus 0.4em\relax ACM, 2017, pp. 42--52. [Online]. Available: \url{https://dl.acm.org/doi/10.1145/2909824.3020207}
\BIBentrySTDinterwordspacing

\bibitem{addleseeMultipartyConversationalSocial2024}
A.~Addlesee, N.~Cherakara, N.~Nelson, D.~Hernández~García, N.~Gunson \emph{et~al.}, ``A {{Multi-party Conversational Social Robot Using LLMs}},'' in \emph{Companion of the 2024 {{ACM}}/{{IEEE International Conference}} on {{Human-Robot Interaction}}}, ser. {{HRI}} '24.\hskip 1em plus 0.5em minus 0.4em\relax Association for Computing Machinery, 2024, pp. 1273--1275.

\bibitem{muraliImprovingMultipartyInteractions2023}
P.~Murali, I.~Steenstra, H.~S. Yun, A.~Shamekhi, and T.~Bickmore, ``Improving {{Multiparty Interactions}} with a {{Robot Using Large Language Models}},'' in \emph{Extended {{Abstracts}} of the 2023 {{CHI Conference}} on {{Human Factors}} in {{Computing Systems}}}.\hskip 1em plus 0.5em minus 0.4em\relax ACM, 2023, pp. 1--8.

\bibitem{addleseeMultipartyMultimodalConversations2024}
A.~Addlesee, N.~Cherakara, N.~Nelson, D.~Hernández~García, N.~Gunson, W.~Sieińska \emph{et~al.}, ``Multi-party {{Multimodal Conversations Between Patients}}, {{Their Companions}}, and a {{Social Robot}} in a {{Hospital Memory Clinic}},'' in \emph{Proceedings of the 18th {{Conference}} of the {{European Chapter}} of the {{Association}} for {{Computational Linguistics}}: {{System Demonstrations}}}, N.~Aletras and O.~De~Clercq, Eds.\hskip 1em plus 0.5em minus 0.4em\relax Association for Computational Linguistics, 2024, pp. 62--70.

\bibitem{bensch2024}
\BIBentryALTinterwordspacing
C.~Bensch, A.~Müller, O.~Chojnowski, and A.~Richert, ``Beyond {{Binary Dialogues}}: {{Research}} and {{Development}} of a {{Linguistically Nuanced Conversation Design}} for {{Social Robots}} in {{Group}}–{{Robot Interactions}},'' \emph{Applied Sciences}, vol.~14, no.~22, p. 10316, 2024. [Online]. Available: \url{https://www.mdpi.com/2076-3417/14/22/10316}
\BIBentrySTDinterwordspacing

\bibitem{gilletTemplatesGraphNeural2025}
S.~Gillet, S.~Thompson, I.~Leite, and M.~Vázquez, ``Templates and {{Graph Neural Networks}} for {{Social Robots Interacting}} in {{Small Groups}} of {{Varying Sizes}},'' in \emph{Proceedings of the 2025 {{ACM}}/{{IEEE International Conference}} on {{Human-Robot Interaction}}}, ser. {{HRI}} '25.\hskip 1em plus 0.5em minus 0.4em\relax IEEE Press, 2025, pp. 458--467.

\bibitem{openaiGPT4TechnicalReport2024}
\BIBentryALTinterwordspacing
OpenAI, J.~Achiam, S.~Adler \emph{et~al.} (2024) {{GPT-4 Technical Report}}. [Online]. Available: \url{http://arxiv.org/abs/2303.08774}
\BIBentrySTDinterwordspacing

\bibitem{dubeyLlama3Herd2024}
\BIBentryALTinterwordspacing
A.~Dubey, A.~Jauhri, A.~Pandey \emph{et~al.} (2024) The {{Llama}} 3 {{Herd}} of {{Models}}. [Online]. Available: \url{http://arxiv.org/abs/2407.21783}
\BIBentrySTDinterwordspacing

\bibitem{guzmanMakingAISafe2017}
A.~L. Guzman, ``Making {{AI Safe}} for {{Humans}}: {{A Conversation}} with {{Siri}},'' in \emph{Socialbots and Their Friends: Digital Media and the Automation of Sociality}, R.~W. Gehl and M.~Bakardjieva, Eds.\hskip 1em plus 0.5em minus 0.4em\relax Routledge, Taylor \& Francis Group, 2017.

\bibitem{bohusModelsMultipartyEngagement2009}
D.~Bohus and E.~Horvitz, ``Models for multiparty engagement in open-world dialog,'' in \emph{Proceedings of the {{SIGDIAL}} 2009 {{Conference}} on {{The}} 10th {{Annual Meeting}} of the {{Special Interest Group}} on {{Discourse}} and {{Dialogue}} - {{SIGDIAL}} '09}.\hskip 1em plus 0.5em minus 0.4em\relax Association for Computational Linguistics, 2009, pp. 225--234.

\bibitem{jungRobotsWildTime2018}
M.~Jung and P.~Hinds, ``Robots in the {{Wild}}: {{A Time}} for {{More Robust Theories}} of {{Human-Robot Interaction}},'' \emph{ACM Transactions on Human-Robot Interaction}, vol.~7, no.~1, pp. 2:1--2:5, 2018.

\bibitem{chojnowskiHumanlikeNonverbalBehavior2025a}
O.~Chojnowski, A.~Eberhard, M.~Schiffmann, A.~Müller, and A.~Richert, ``Human-like {{Nonverbal Behavior}} with {{MetaHumans}} in {{Real-World Interaction Studies}}: {{An Architecture Using Generative Methods}} and {{Motion Capture}},'' in \emph{Proceedings of the 2025 {{ACM}}/{{IEEE International Conference}} on {{Human-Robot Interaction}}}, ser. {{HRI}} '25.\hskip 1em plus 0.5em minus 0.4em\relax IEEE Press, 2025, pp. 1279--1283.

\bibitem{jocherUltralyticsYolov5V312020}
\BIBentryALTinterwordspacing
G.~Jocher, A.~Stoken, J.~Borovec \emph{et~al.}, ``Ultralytics/yolov5: V3.1,'' Zenodo, 2020. [Online]. Available: \url{https://zenodo.org/records/4154370}
\BIBentrySTDinterwordspacing

\bibitem{googleaiedgeMediaPipe}
\BIBentryALTinterwordspacing
{Google AI Edge}, ``{{MediaPipe}},'' Google. [Online]. Available: \url{https://github.com/google-ai-edge/mediapipe}
\BIBentrySTDinterwordspacing

\bibitem{fineGroupCultureInteraction2012}
G.~A. Fine, ``Group {{Culture}} and the {{Interaction Order}}: {{Local Sociology}} on the {{Meso-Level}},'' \emph{Annual Review of Sociology}, vol.~38, pp. 159--179, 2012.

\bibitem{fieldDiscoveringStatisticsUsing2024}
A.~Field, \emph{Discovering Statistics Using IBM SPSS Statistics}, 6th~ed.\hskip 1em plus 0.5em minus 0.4em\relax Sage Publications Ltd., 2024.

\bibitem{faulStatisticalPowerAnalyses2009}
F.~Faul, E.~Erdfelder, A.~Buchner, and A.-G. Lang, ``Statistical power analyses using {{G}}*{{Power}} 3.1: {{Tests}} for correlation and regression analyses,'' \emph{Behavior Research Methods}, vol.~41, no.~4, pp. 1149--1160, 2009.

\end{thebibliography}

\end{document}